\documentclass[pdflatex,sn-mathphys-num]{sn-jnl}

\usepackage{graphicx}%
\usepackage{multirow}%
\usepackage{amsmath,amssymb,amsfonts}%
\usepackage{amsthm}%
\usepackage{mathrsfs}%
\usepackage[title]{appendix}%
\usepackage{xcolor}%
\usepackage{textcomp}%
\usepackage{manyfoot}%
\usepackage{booktabs}%
\usepackage{algorithm}%
\usepackage{algorithmicx}%
\usepackage{algpseudocode}%
\usepackage{listings}%

\theoremstyle{thmstyleone}%
\theoremstyle{thmstyletwo}%

\theoremstyle{thmstylethree}%

\begin{document}

\title[Article Title]{Synthetic Lung X-ray Generation through Cross-Attention and Affinity Transformation}


\author*[1,2]{\fnm{Ruochen} \sur{Pi}}\email{pirch2024@shanghaitech.edu.cn}

\author[3]{\fnm{Lianlei} \sur{Shan}}\email{shanlianlei18@mails.ucas.edu.cn}

\affil[1]{\orgdiv{Information Hub}, \orgname{Hong Kong University of Science and Technology(Guangzhou)}, \orgaddress{\city{Guangzhou}, \postcode{511453}, \state{Guangdong}, \country{China}}}

\affil[2]{\orgdiv{Faculty of Engineering}, \orgname{University of Sydney}, \orgaddress{\city{Sydney}, \postcode{2006}, \state{NSW}, \country{Australia}}}

\affil[3]{\orgdiv{School of Computer Science and Technology}, \orgname{University of Chinese Academy of Sciences}, \orgaddress{ \city{Beijing}, \postcode{100864}, \country{China}}}




\abstract{Collecting and annotating medical images is a time-consuming and resource-intensive task. However, generating synthetic data through models such as Diffusion offers a cost-effective alternative. This paper introduces a new method for the automatic generation of accurate semantic masks from synthetic lung X-ray images based on a stable diffusion model trained on text-image pairs. This method uses cross-attention mapping between text and image to extend text-driven image synthesis to semantic mask generation. It employs text-guided cross-attention information to identify specific areas in an image and combines this with innovative techniques to produce high-resolution, class-differentiated pixel masks. This approach significantly reduces the costs associated with data collection and annotation. The experimental results demonstrate that segmentation models trained on synthetic data generated using the method are comparable to, and in some cases even better than, models trained on real datasets. This shows the effectiveness of the method and its potential to revolutionize medical image analysis.}

\keywords{Image segmentation, Image annotation, Medical imaging, data generation}



\maketitle

\section{Introduction}\label{sec1}

\subsection{Significance of the Research Topic}

Medical image segmentation, essential for identifying anatomical structures in medical imaging, is critical for improving diagnosis, treatment planning, and clinical research. This task is challenged by medical images' complex and noisy nature, necessitating automatic segmentation methods to bypass the costly, time-consuming, and error-prone manual segmentation process. Furthermore, developing and refining automatic segmentation methods demand extensive annotated datasets, which are difficult and laborious to prepare. The acquisition of medical image data is often fraught with challenges, including privacy concerns and the scarcity of available datasets. These factors make collecting and annotating large volumes of medical image data a formidable task. Despite these hurdles, recent advances in deep learning have significantly improved the accuracy and efficiency of medical segmentation, establishing it as an increasingly vital research area with profound implications for patient care and medical research. As the field progresses, it promises to expand the capabilities of medical imaging analysis, albeit with the ongoing challenge of data acquisition and annotation~\cite{Wang2022MedicalIS}.

The variability in medical images, stemming from various imaging technologies like MRI, CT, and PET, introduces additional complexities such as partial volume effects, noise, and insufficient resolution. These challenges are further amplified by the complex anatomy of human organs, which vary in shape and intensity, and the blurred boundaries between adjacent organs~\cite{Alzahrani2021BiomedicalIS}~\cite{Liu2021ReviewDL}.

\subsection{Previous Methods and Their Limitations}

Medical image segmentation methods can separated into three main categories: region-based, deformable models, and machine learning approaches, each with distinct techniques and limitations. 
\begin{itemize}
    \item \textbf{Region-based methods}, including edge detection and thresholding, are simple and fast but struggle with noise and poor contrast, making them less effective for ultrasound and more suited for high-contrast images like X-rays. Techniques like region growing offer less noise sensitivity but require user interaction, while Markov Random Fields and watershed methods can be complex or prone to over-segmentation.
    \item \textbf{Deformable models}, such as level set methods and Active Contour Models, are robust across various image types, especially for 3D magnetic resonance imaging, but require significant computation and are sensitive to initialization.
    \item \textbf{Learning-based methods}, in image segmentation encompass supervised, unsupervised, and semi-supervised approaches, each differing in applicability and effectiveness. Supervised learning methods, such as Convolutional Neural Networks (CNNs) and their variants like VGG16, GoogleNet, and ResUNet, are highly reliable for feature extraction and pixel-level segmentation, though they often require substantial preprocessing or post-processing and can suffer from overfitting, underfitting, and large dataset requirements, raising privacy concerns in medical data. Unsupervised techniques, including K-means and Fuzzy C-means, are effective for simple segmentation tasks but may struggle with noisy images due to their reliance on less complex models. Semi-supervised methods offer a middle ground, leveraging a smaller amount of labelled data augmented with larger unlabeled datasets to improve learning efficiency and robustness, particularly in environments with noisy or incomplete data.
\end{itemize}

\subsection{Improvements of Our Method Over Previous Ones}

Our approach, termed DiffMask, revolutionizes medical image segmentation by eliminating the need for pixel-level annotations, leveraging zero-shot text-to-image models like Stable Diffusion, trained on vast image-text pair corpora. It introduces two innovations. First, adaptive thresholding converts attention maps into precise binary masks. Second, bridging the domain gap with retrieval-based prompts and data augmentation, like stitching, to enhance data diversity, which allows DiffMask to generate limitless annotated images for any category, facilitating the training of segmentation models without manual data labelling.

\subsection{Summary of Contributions}

\begin{itemize}
    \item \textbf{Fusing Text and Image via Cross-Attention Maps: }This method automates the generation of semantic masks from text-to-image diffusion model attention maps, blending textual and visual data for image generation closely aligned with text prompts.

    \item \textbf{Synthetic Image and Mask Creation with Text-Supervised Models:} Utilizes text-supervised pretrained diffusion models to automatically create contextually relevant synthetic images and their masks, eliminating manual annotation.

    \item \textbf{Zero-shot Learning Capability:} Enables the generation of images and accurate masks for unseen classes, greatly broadening the model's utility and flexibility across various domains without additional specific training data.
\end{itemize}

\section{Related Work}\label{sec2}

\subsection{Medical Image Segmentation}

Medical image segmentation methods are categorized into region-based, deformable models, learning-based methods, and other approaches, as mentioned in the previous part, focusing on deep learning within the learning-based category.

In 2015, Long, Shelhamer, and others introduced the Fully Convolutional Network (FCN)~\cite{FCNLong2015FullyCN}, which extended traditional Convolutional Neural Networks (CNNs) to pixel-level classification, achieving end-to-end image segmentation. Upsampling and skip connections improved segmentation accuracy but might lack precision in handling detailed information. Ronneberger, Fischer, and Brox proposed UNet~\cite{UNETAlzahrani2021BiomedicalImageSegmentation}, specially designed for medical image segmentation, featuring a symmetric structure and numerous skip connections to combine low-level and high-level features effectively. The UNet architecture can achieve high-precision segmentation with a small amount of data but might not be robust enough for very complex or non-standard images. Milletari and colleagues introduced V-Net~\cite{Milletari2016VNet}, a network structure specifically designed for 3D medical imaging. It employs 3D convolutions and the Dice coefficient loss function to improve the segmentation of 3D images, but it requires significant computational resources and strong hardware support due to its large computational demand. SegNet~\cite{Badrinarayanan2017SegNet}, developed by Vijay Badrinarayanan, Alex Kendall, and Roberto Cipolla from the University of Cambridge, is a convolutional neural network architecture for deep semantic segmentation, particularly suited for road scene understanding and indoor object segmentation. It uses an encoder-decoder architecture and employs upsampling in the decoder, reducing the model's parameter count by remembering pooling indices from the encoder. However, it might not be as precise as other more advanced models in processing complex scenes with rich details. DeepLab~\cite{DEEPLABChen2016DeepLab}, developed by Google's research team including Liang-Chieh Chen and others, is a series of models for deep learning semantic segmentation that incorporates various advanced technologies. Atrous convolution (or dilated convolution) was introduced to expand the receptive field without reducing image resolution. Fully connected CRFs (Conditional Random Fields) were used as a post-processing step to refine segmentation edges. DeepLabV3 and DeepLabV3+ introduced atrous spatial pyramid pooling to enhance segmentation accuracy further. However, DeepLab requires a relatively large computational load, especially when using atrous convolution and CRFs. The model may need specific adjustments for certain types of images to improve accuracy.

\subsection{Image Generation}

Image generation is a basic and challenging task in computer vision~\cite{DiffuMaskWu2023DiffuMask}. There are several mainstream methods for the task, including Generative Adversarial Networks (GAN) ~\cite{GANsgoodfellow2014generative}, Variational autoencoders (VAE)~\cite{VAEskingma2013auto}, flow-based models~\cite{Followdinh2016density}, and Diffusion Probabilistic Models (DM) ~\cite{DMsohl2015deep}.

Goodfellow et al.~\cite{GANsgoodfellow2014generative} introduced a GAN, a framework composed of a generator and a discriminator, designed for the generation of high-quality images. The generator endeavors to produce images that appear authentic, while the discriminator's task is to differentiate between real images and those generated by the generator. GANs have demonstrated their formidable capabilities across multiple domains, including image synthesis and style transfer. However, the stability of GAN training and the issue of mode collapse (resulting in repetitive or meaningless outputs) represent significant drawbacks of this technology. Brock et al. ~\cite{BigGANBrock2018LargeSG}achieved higher quality image generation by incorporating larger training datasets and increasing the model size within the GAN architecture, a method termed BigGAN. BigGAN demonstrates that scaling up the model and training data can significantly enhance the quality of generated images. However, this approach also incurs higher training costs and demands substantial computational resources. Karras et al.~\cite{StyleGANKarras2019StyleGAN} introduced a novel generative architecture, StyleGAN, capable of manipulating high-level attributes and style transitions in generated images. By adjusting the inputs at different layers, allows for precise control over various features of the image, such as facial characteristics and hairstyles. StyleGAN has shown exceptional performance in the domain of facial generation. However, its complex training procedure and the high computational cost have limited its broader application. In recent years, there have also been many breakthrough achievements for Zero-shot. Couairon et al.~\cite{Couairon2023ZeroShotSL} introduced ZestGuide, a zero-shot segmentation guidance method that can be integrated into pre-trained text-to-image diffusion models without any additional training. It leverages implicit segmentation maps extracted from cross-attention layers, aligning the generation with input masks. Experimental results demonstrated a combination of high image quality and accurate alignment with input segmentation, marking quantitative and qualitative improvements over previous works.

\subsection{Label Efficiency}
To mitigate annotation costs, various methods can be explored, including interactive human-in-the-loop annotation, nearest neighbour mask transfer, or supervision with weak/inexpensive mask annotations, as well as synthetic data generation. Techniques such as interactive methods and nearest-neighbour transfer leverage the predictive capabilities of machine learning models to reduce redundant work and improve annotation efficiency. Meanwhile, to reduce costs, some methods, like image-level labelling or scribble annotation, compromise on annotation precision. Although these approaches are less costly, they might lead to diminished model performance. In contrast, synthetic data offers numerous advantages, including reducing data costs without the need for image collection and offering limitless availability to enhance data diversity~\cite{DiffuMaskWu2023DiffuMask}.

Acuna et al.~\cite{Acuna2018Efficient} introduced an efficient interactive annotation with Polygon-RNN++, which improves interactive object annotation in images by utilizing a Graph Neural Network, allowing accurate high-resolution object annotations with reduced human interaction. Zhao et al.~\cite{Zhao2020Weakly} use weakly supervised cell Segmentation, which introduces weakly supervised training schemes for cell segmentation using only a single point annotation per cell, exploring self-training, co-training, and hybrid-training schemes to improve annotation efficiency. Saisho et al.~\cite{Saisho2021Human} also proposes an interactive data annotation method that combines weakly supervised learning with uncertainty-based active learning strategies to reduce annotation costs across various fields. Active Learning with Contrastive Explanations (ALICE)~\cite{Liang2020ALICE:} utilizes contrastive natural language explanations within an expert-in-the-loop framework to enhance data efficiency in visual recognition tasks, achieving performance gains comparable to using significantly more labelled training data.

\section{Methodology}

\subsection{Overview of the Method}
\begin{figure*}[!t]
    \centering
    \includegraphics[width=0.8\textwidth]{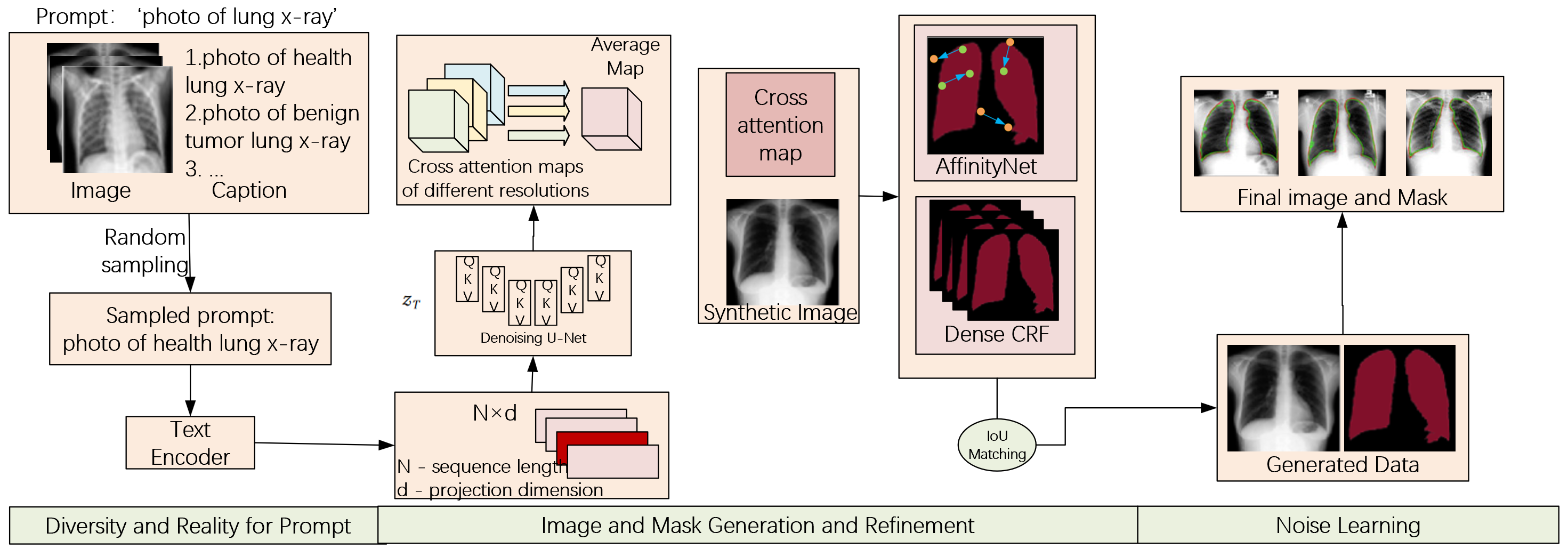} 
    \caption{\textbf{Overview of the Method}: The workflow of our method, highlights the integration of cross-attention mechanisms in generative models to bridge synthetic and real data for seamless mask generation and segmentation training.}
    \label{fig:fig1}
\end{figure*}
To train segmentation methods with synthetic data for application on real images, we leverage cross-attention in generative models to bridge the gap between synthetic and real data, offering novel insights and solutions through a three-step process, which includes text-to-image generation, mask adaptation and refinement, and segmentation network training. 
This paper leverages the cross-attention maps between text and images within diffusion models to synthesize images and their corresponding high-quality semantic masks without the need for manual intervention. This process involves extracting specific category/word regions from these attention maps and integrating them with techniques to create detailed category-distinguishing masks. In refining these masks, the method includes using adaptive thresholding to convert attention maps into binary masks and employing AffinityNet\cite{Affinityfrey2007clustering} and DenseCRF\cite{DenseCRFkrahenbuhl2011efficient} for further refinement to enhance mask precision, especially in capturing fine details. This approach significantly reduces the workload and cost associated with collecting and annotating data for semantic segmentation tasks. It is shown in Fig.\ref{fig:fig1}.

\subsection{Text-to-Image Generation}

In this study, the text-to-image process employs a cross-attention mechanism to merge text descriptions with image generation, ensuring images match the textual content, which aids in generating data for tasks like semantic segmentation that need precise understanding. Starting with random Gaussian noise, the process uses text prompts, a text encoder, a VAE, and a U-shaped net. The mechanism projects noise image features into query vectors and text prompts into key and value matrices, creating cross-attention maps. These maps blend visual and textual data, updating the image's spatial features to align with text prompts.

The noise image's visual features, $\phi(z_t)$, in space $H \times W \times C$, are flattened and linearly projected into query vectors $Q = \ell_Q(\phi(z_t))$. The text prompt $\mathcal{P}$ is projected into the text embedding \(\tau_\theta (\mathcal{P} ) \in \mathbb{R} ^{N \times d}\), where $N$ is the sequence length of text tokens, and $d$ is the dimensionality of the latent projection, then mapped to the key matrix $K = \ell_K(\tau_\theta(\mathcal{P}))$, and the value matrix $V = \ell_V(\tau_\theta (\mathcal{P}))$. The cross-attention map can be presented as follows:

\begin{equation}
A=\mathrm{{Softmax}}\left({\frac{Q K^{T}}{\sqrt{d}}}\right)
\end{equation}
where \(A\) representing a tensor reshaped into dimensions \(\mathbb{R}^{H \times W \times N}\). Considering the \(j\)-th element of the text sequence, one can derive the associated weight matrix \(A_j\) within the space \(\mathbb{R}^{H \times W}\) mapped by \(\phi(\mathbf{z}_t)\). Subsequently, the interaction attention outcome is denoted by \(\hat{\phi}(\mathbf{z}_t) = AV\), and this is utilized for the refinement of spatial feature maps \(\phi(\mathbf{z}_t)\).

\subsection{Mask Adaptation and Refinement}
In this study, we employed an innovative mask refinement method to generate preliminary masks through cross-attention maps and further enhance mask quality via DenseCRF and Affinity. It explains the methodology for obtaining average cross-attention maps from various layers of a UNet, across different resolutions, and how these maps are aggregated to create a single, informative attention map. The section further elaborates on converting these attention maps into binary maps through standard binarization and adaptive threshold techniques, optimizing the mask quality for semantic segmentation tasks.
   \begin{figure}[H]
       \centering
       \includegraphics[width=0.75\linewidth]{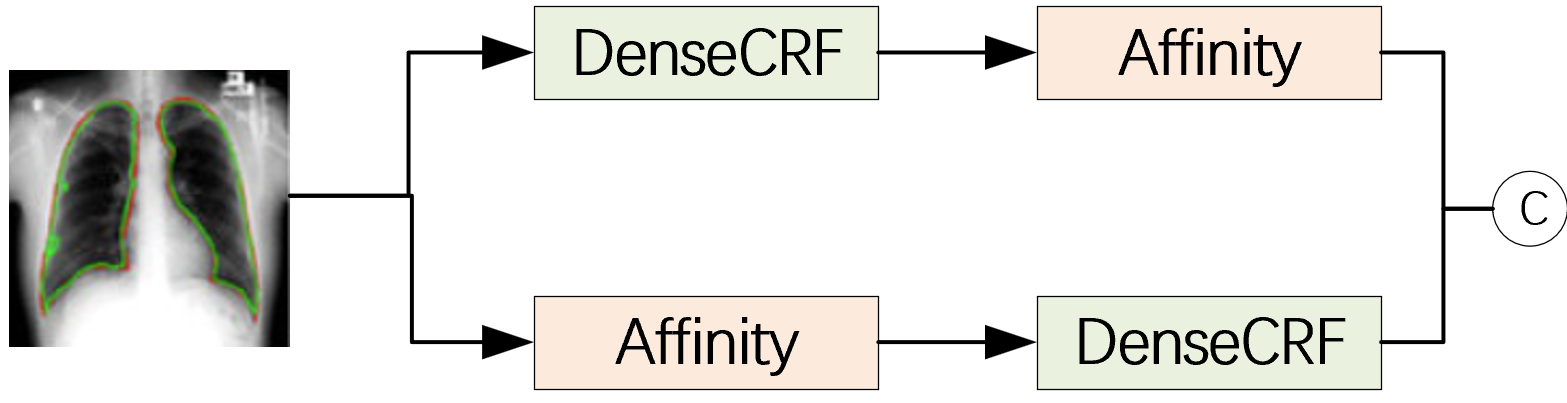}
       \caption{\textbf{Method 1}: Cross-fusion of DenseCRF and Affinity modules synergistically refines mask quality by combining spatial coherence with semantic optimization.}
       \label{fig:fig2}
   \end{figure}
   \begin{figure}[H]
       \centering
       \includegraphics[width=0.75\linewidth]{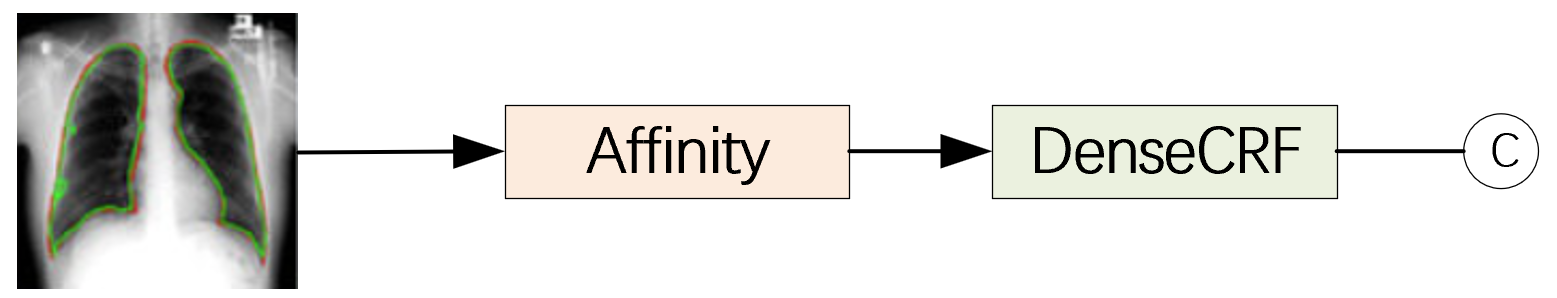}
       \caption{\textbf{Method 2}: Sequential application of Affinity for semantic drafting followed by DenseCRF for spatial and visual refinement achieves a precise and polished final mask.}
       \label{fig:fig3}
   \end{figure}
To obtain a more accurate mask, two different methods are compared. One is the cross-fusion of DenseCRF and Affinity, the other is the sequential use of Affinity and DenseCRP.
    
Method 1, the image is initially received and then separately processed by the DenseCRF and Affinity modules for preliminary mask refinement. DenseCRF is primarily responsible for enhancing the spatial coherence and edge clarity of the mask, while the Affinity module optimizes the mask using the semantic affinity between the image and text to select the best binarization threshold adaptively. Subsequently, the two modules are used in a cross-application manner, the Affinity module further refines the mask processed by DenseCRF. Similarly, the mask processed by Affinity is optimized spatially by DenseCRF. This cross-utilization aims to combine the advantages of both methods to improve mask quality further. Ultimately, the best result from both processing streams is selected as the final output.

Method 2 begins with applying the Affinity module, which uses an adaptive threshold to generate a rough mask aiming to capture image regions semantically related to the input text. The focus here is on determining which parts of the image belong to the foreground based on semantic information between the image and text. The processed mask is then passed to the DenseCRF module, further refining the mask by improving its spatial coherence and visual quality. This step considers the colour and spatial information between pixels, aiming to make the mask smoother and more accurate. The continuous processing by Affinity and DenseCRF ultimately produces a mask that is optimized both semantically and visually, serving as the final output.

Both AffinityNet and DenseCRF contribute significantly to the overall quality and accuracy of the generated masks, each serving a distinct role in the mask refinement process. AffinityNet optimizes the selection of thresholds for binarization based on semantic affinity, while DenseCRF enhances the spatial coherence and visual quality of the resultant masks. This comprehensive approach to mask generation and refinement leverages the detailed semantic information captured by cross-attention mechanisms in diffusion models, further enhanced by adaptive thresholding techniques for improved mask quality. Through the integration of advanced binarization processes and semantic affinity learning, the method achieves precise segmentation that aligns closely with the textual prompts, facilitating more accurate and contextually relevant image manipulation and generation tasks. 

\textbf{Cross-Attention Map Generation Refinement}Utilizing the softmax function, the cross-attention map is derived from the interaction between the text (query) and the image (key) features. This process is conducted across different layers of a UNet model, corresponding to multiple resolutions ($8\times8, 16 \times16, 32 \times 32, 64 \times64$) and diffusion steps, to capture detailed semantic information related to the text prompt. A comprehensive attention map is generated by averaging attention maps from multiple layers and time steps. This averaged map represents the combined attention across all considered factors, providing a more robust representation of the relevant features for mask generation.
\begin{equation}
\hat{\mathcal{A}}_{j}=\frac{1}{S\cdot T}\sum_{s\in S,t\in T}\frac{\mathcal{A}_{j}^{s,t}}{\mathrm{max}(\mathcal{A}_{j}^{s,t})}
\end{equation}
where  \(\hat{A}_j\) is average cross-attention map for the \(j\)-th text token, \(S\) is total number of steps in the diffusion process, \(T\) is total number of layers in the UNet architecture, \(s\) is a specific step in the diffusion process, \(t\) is a specific layer in the UNet architecture. It can get the \(A_{st}^j\), the cross-attention map for the \(j\)-th text token at step \(s\) and layer \(t\), then \(\max(A_{st}^j)\) is normalization term, which is the maximum value of the attention map for the \(j\)-th text token at step \(s\) and layer \(t\), used to normalize the attention values to ensure they are between 0 and 1.
This equation normalizes and averages the attention maps across all specified layers and steps, producing a final, averaged attention map that highlights areas of the image that are most relevant to the \(j\)-th text token. This is useful for tasks such as generating detailed segmentation masks from textual descriptions using diffusion models.

\textbf{Standard Binarization (DenseCRF)} 
Utilizing cross-attention maps, we first generate an average attention map that reflects the relevance between textual descriptions and image regions. By converting this average attention map into a binary image, we define the foreground areas. To further refine the binary mask generated by the binarization process, a Dense Conditional Random Field (DenseCRF) is used. It is based on the principle of considering colour and pixel proximity to enhance the local consistency and edge detail of the segmentation mask. The DenseCRF input is a binary mask generated after applying the threshold to the probability graph (note the graph). It refines the binary mask by considering local relationships between adjacent pixels. Specifically, DenseCRF adjusts the mask boundaries based on the colour similarity and distance of adjacent pixels. This ensures that contiguous areas that look similar are more likely to be labelled consistently, helping to reduce jagged edges and small isolated areas in the mask that do not correspond to separate objects. The result is a smoother, more accurate mask that better conforms to the shapes and boundaries of objects within the image, resulting in improved segmentation quality.
\begin{equation}
B=\mathrm{DenseCRF}\Big(\left[\gamma;\hat{A}_{j}\right]_{\mathrm{argmax}}\Big)\ 
\end{equation}
where \(\hat{A}_j \) is the average attention map for the j-th text token, where attention maps from different layers and times are aggregated. The resulting \(B\) represents the binary graph generated after the \(\gamma \) threshold processing is applied. In this case, a binary graph is an image in which each pixel is assigned a value of 0 or 1, representing the background and foreground, respectively.

\textbf{Semantic Affinity Learning}
AffinityNet significantly contributes to enhancing the segmentation mask by assessing the semantic affinity between pixels. This solves the challenge of medium confidence score pixels that do not strongly indicate the foreground or the background. These pixels present uncertainty in mask generation and significantly affect the quality and accuracy of segmentation. AffinityNet is designed to evaluate the semantic affinity between adjacent pixel pairs, focusing on pixels with moderate confidence scores. During the training phase, pixels in the middle score range are considered neutral. The network ignores pairs involving neutral pixels to avoid ambiguity. The affinity between two coordinates is determined based on their semantic similarity. If they belong to the same class, their affinity label is set to 1, indicating a positive pair; Otherwise, the value is set to 0, indicating a negative pair. This approach helps refine the mask by providing a more nuanced understanding of which pixels should be grouped based on semantic relationships inferred by AffinityNet. Through this process, we get $\hat{B}$, which is a coarse affinity map. AffinityNet can predict it for each class of each image.

\textbf{Fusion Threshold Adjustment}
Considering the significant shape and area differences between different images and categories, this study introduces a fusion threshold adjustment strategy aimed at dynamically determining the optimal binarization threshold. We leverage AffinityNet to assess the category affiliation of medium confidence pixels, thereby globally optimizing the semantic mask threshold that represents the overall prototype. The formula to find the optimal threshold is as follows:
\begin{equation}
\hat{\gamma}=\arg\operatorname*{max}_{\gamma\in\Omega}\sum{ L}_{\operatorname*{match}}(\hat{B},B_{\gamma})
\end{equation}
where ${ L}_{\operatorname*{match}}$ calculate the cost of matching $\hat{B}$ and $B_{\gamma}$ pairs, and $\gamma\in\Omega$, and $\Omega = \left \{ \gamma_i \right \} ^L_{i=1}$.

\subsection{Segmentation Networks Training}
In the experiments, advanced segmentation models such as UNet, Attention-UNet, TransUNet, and Mask2former were selected as the foundation for training. These models were utilized to learn the mapping from generated images to their corresponding masks, thereby simulating the segmentation tasks of real-world images. The training of the generated dataset was conducted using cross-validation. This means the dataset was divided into multiple parts, with each part taking turns serving as the test set while the others were used as the training set. This method enhances the model's generalization ability and ensures the reliability of the evaluation results. Unlike traditional methods that mix real and generated images for training, this approach first trains the backbone with generated data and then refines the outcomes with real images. The backbone includes ResNet50, ResNet101, and ViT.
\begin{figure}[H]
    \centering
    \includegraphics[width=0.7\linewidth]{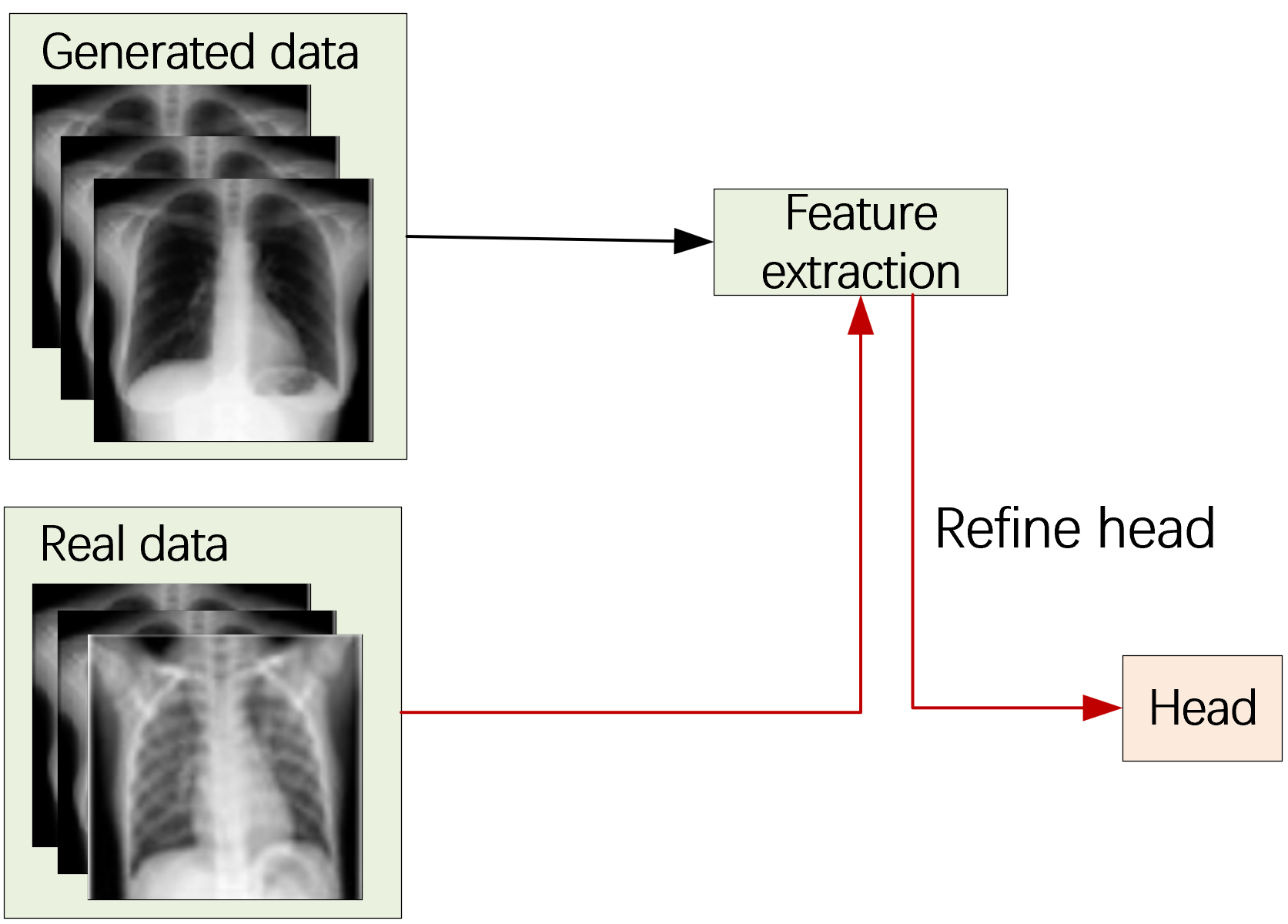}
    \caption{\textbf{Segmentation Networks Training}: The diagram shows a process where features from generated data and refinements from real data are integrated to enhance model output.}
    \label{fig:fig4}
\end{figure}

\section{Experiment}
\subsection{Data and Introduction}
The experiments utilized real data from a dataset available in a Kaggle competition, which contains 3520 manually annotated lung images with their corresponding labels. A selection process was undertaken to curate the data, combining both synthetic and real images for the training dataset, while the test dataset was comprised of 110 real images. The performance improvement of different models was evaluated based on the Intersection of Union (IoU) values with the real data. 
Training was conducted on various models using the training dataset (a mix of synthetic and real data). The trained models were then applied to segment the real images in the test dataset. The segmentation results were compared with the actual segmentation maps to calculate the IoU values, assessing the impact of synthetic data on the performance of different segmentation models. This evaluation method aims to determine the effectiveness of using synthetic data in enhancing the model's ability to generalize and accurately perform segmentation tasks on real-world images.

\subsection{Implementation Details}
In this study, we integrate pre-trained components such as Stable Diffusion, text encoder, AffinityNet, and DenseCRF, using them without fine-tuning Stable Diffusion. Specifically, we train AffinityNet and DenseCRF for each category identified within Stable Diffusion. The training process adheres to parameter optimizations and configurations (like initialization, data augmentation, batch size, and learning rate) as detailed in their respective original publications. We generate synthetic data for each category of medical images, producing 10,000 images per category. Through noise learning, set with an \(\alpha\) of 0.7, we filter this down to 3,000 images per category, resulting in 60,000 images across 20 categories for the final training set, all at a spatial resolution of \(512 \times 512\). The study focuses on the segmentation of single objects within images, avoiding multiclass segmentation due to potential instability and quality issues tied to the generative capabilities of Stable Diffusion. For evaluation, we employ a variety of UNet models, including UNet, Attention-UNet, TransUNet, and Mask2Former, as baselines. All experiments leverage the computational capabilities of eight NVIDIA GeForce RTX 3090 GPUs. This setup is designed to thoroughly evaluate how synthetic data, generated via sophisticated modeling techniques, can enhance the performance of segmentation models in a range of medical imaging tasks.

\subsection{Evaluation metrics}
In various segmentation models, the training set utilizes synthetic data to train the models, leading to models that perform segmentation on real images. The performance of these models is assessed by calculating the Intersection over Union (IoU) value between the model-generated binary map and the actual binary map. The IoU value is calculated using the following formula:

\begin{equation}{\mathrm{IoU}}={\frac{\mathrm{Area~of~Overlap}}{\mathrm{Area~of~Union}}}\end{equation}
where the Area of Overlap represents the area of the intersection between the predicted binary map and the actual binary map, and the Area of Union represents the total area encompassed by both the predicted and actual binary maps combined.

This is specifically achieved by calculating the ratio between the intersection area and the union area of the two binary maps. A higher IoU value indicates that the binary map segmented by the model is more similar to the actual binary map, suggesting good segmentation performance by the model and implying that synthetic data can enhance model performance. Conversely, a lower IoU value indicates poor segmentation performance by the model, suggesting that synthetic data does not improve model performance.

\subsection{Experiment Results and Analysis}

We analyzed the impact of generated images on the performance of UNet and TransUnet models, and found that the introduction of 5000 generated images significantly improves the IoU values of both models,as shown in Table \ref{tab:tab1} and Table \ref{tab:tab2}.

\begin{table}[h]
\centering
\caption{IoU (\%) of UNet with Varying Real/Generated Proportions}
\begin{tabular}{lllll}
\toprule
Method & Backbone & Real Img & Generate Img & IoU \\
\midrule
    
Unet     & ResNet-50       & 1   & 0   & 17.1  \\
Unet     & ResNet-50       & 1   & 5000   & 88.7  \\
\hline
Unet     & ResNet-50       & 3   & 0   & 22.4  \\
Unet     & ResNet-50       & 3   & 5000   & 90.2  \\
\hline
Unet     & ResNet-50       & 5   & 0   & 45.3  \\
Unet     & ResNet-50       & 5   & 5000   & 91.0  \\
Unet     & ResNet-50       & All   & 0   & 84.4  \\
Unet     & ResNet-50       & All   & 5000   & 94.2  \\
\bottomrule
\end{tabular}

\label{tab:tab1}
\end{table}

\begin{table}[h]
\centering
\caption{IoU (\%) of TransNet with Varying Real/Generated Proportions}
\begin{tabular}{lllll}
\toprule
Method & Backbone & Real Img & Generate Img & IoU(\%) \\
\midrule
    
TransUnet     & ResNet-50       & 1   & 0   & 8.1  \\
TransUnet     & ResNet-50       & 1   & 5000   & 91.7  \\
\hline
TransUnet     & ResNet-50       & 3   & 0   & 11.0  \\
TransUnet     & ResNet-50       & 3   & 5000   & 91.9  \\
\hline
TransUnet     & ResNet-50       & 5   & 0   & 25.3  \\
TransUnet     & ResNet-50       & 5   & 5000   & 91.9  \\
TransUnet     & ResNet-50       & All   & 0   & 87.4  \\
TransUnet     & ResNet-50       & All   & 5000   & 94.9  \\
\bottomrule
\end{tabular}

\label{tab:tab2}
\end{table}



For the UNet model, the IoU increased dramatically from 17.1 to 88.7 with the addition of 5000 generated images, even when only one real image was used. This trend continued with three and five real images, where the IoU values rose from 22.4 to 90.2 and from 45.3 to 91.0, respectively. When using all available real images, the IoU further improved from 84.4 to 94.2 with the incorporation of synthetic data. Similarly, the TransUnet model exhibited substantial performance gains with synthetic data. The IoU increased from 8.1 to 91.7 with one real image and 5000 generated images. With three and five real images, the IoU values improved from 11.0 to 91.9 and from 25.3 to 91.9, respectively. When all real images were used, the IoU increased from 87.4 to 94.9 with the addition of synthetic data.

These results indicate that synthetic data significantly enhances the performance of both UNet and TransUnet models. However, TransUnet outperforms UNet when ample real data is available, achieving higher IoU values overall. Conversely, UNet shows relatively better performance with limited real data, though it also benefits considerably from synthetic data. This suggests that while both models benefit from synthetic data augmentation, TransUnet is more effective when the data volume is sufficient, whereas UNet can leverage synthetic data to mitigate the impact of data scarcity.

\begin{table}[h]
  \centering
  \caption{IoU (\%) of Different Model Variants Using Synthetic Data}
  \begin{tabular}{lllll}
    \toprule
    Method & Backbone & Real Img & Generate Img & IoU(\%) \\
    \midrule
    
    TransUnet     & ViT       & All   & 0   & 78.3  \\
    TransUnet     & ViT       & All   & 5000   & 97.3  \\
    \midrule
    TransUnet     & ResNet-101       & All   & 0   & 85.9 \\
    TransUnet     & ResNet-101       & All   & 5000   & 95.2  \\
    \midrule
    Attention-Unet     & ResNet-101       & All   & 0   & 84.7  \\
    Attention-Unet     & ResNet-101       & All   & 5000   & 97.8  \\
    \midrule
    Mask2former& ResNet-101       & All   & 0   & 80.2  \\
    Mask2former& ResNet-101       & All   & 5000   & 92.1  \\
    \midrule
    Mask2former& ResNet-50       & All   & 0   & 81.5 \\
    Mask2former& ResNet-50       & All   & 5000   & 96.3  \\
    \bottomrule
  \end{tabular}

  \label{tab:tab3}
\end{table}

The performance of different model variants and their backbone networks after the introduction of 5000 generated images is examined, as shown in Table \ref{tab:tab3}. We found that all models improved in performance, especially TransUnet ViT with the most tips of 97.3 and Attention-Unet ResNet-101 with the highest IoU values and 97.8, highlighting the critical role of synthetic data in improving model accuracy. This effect is partly due to the poor performance of ViT models in the absence of large amounts of training data, a problem mitigated by the addition of generated images.

\subsection{Ablation Study}

We studied the performance variation of the TransUnet ViT model under different numbers of generated images (from 500 to 5000), as shown in Table \ref{tab:tab4}.

\begin{table}[h]
  \centering
  \caption{The IoU (\%) of TransUnet ViT Across Different Image Volumes}
  \begin{tabular}{lllll}
    \toprule
    Method & Backbone & Real Img & Generate Img & IoU(\%) \\
    \midrule
    
    TransUnet     & ViT       & All   & 0   & 78.3  \\
    TransUnet     & ViT       & All   & 500   & 80.3  \\
    TransUnet     & ViT       & All   & 1000   & 87.7  \\
    TransUnet     & ViT       & All   & 1500   & 94.5  \\
    TransUnet     & ViT       & All   & 2000   & 95.8  \\
    TransUnet     & ViT       & All   & 5000   & 97.3  \\
    
    \bottomrule
  \end{tabular}
  
  \label{tab:tab4}
\end{table}


The experimental results demonstrate that increasing the number of generated images significantly enhances the IoU value of the TransUnet ViT model, rising from 78.3 with no generated images to 97.3 with 5000 generated images. The most substantial improvement occurs between 500 and 1500 generated images, where the IoU increases from 80.3 to 94.5. This suggests that the addition of synthetic data within this range provides the model with valuable training examples, enhancing its performance. However, beyond 1500 generated images, the rate of improvement slows. The IoU value increases marginally from 94.5 to 95.8 with 2000 generated images and reaches 97.3 with 5000 generated images. This plateau effect indicates that the model approaches its capacity limit, where additional synthetic data yields diminishing returns. This suggests that while more synthetic data can still improve performance, the impact is less pronounced as the model saturates its learning capability. These findings underscore the importance of both the quantity and quality of training data. While increasing the volume of synthetic data can significantly boost model performance, it is also crucial to ensure that the data is diverse and representative to maximize its effectiveness. Future research could focus on optimizing synthetic data generation processes and exploring advanced model architectures to better leverage large datasets.

In conclusion, synthetic data plays a crucial role in enhancing model performance, particularly when real data is limited. Strategic use of synthetic data can achieve substantial improvements, making it a valuable resource for training robust neural networks.

\subsection{Theoretical Insights}
In the experimental results of our semantic segmentation task, an intriguing phenomenon was observed: neural networks trained on synthetic image data outperformed those trained solely on real-world imagery, as evidenced by a higher accuracy rate. 

Synthetic data is artificially generated data that mimics the characteristics of real-world data. It is used to augment training datasets, particularly when real data is scarce, expensive to obtain, or challenging to label accurately. Synthetic data generation techniques include simulations, computer-generated imagery, and advanced generative models like GANs (Generative Adversarial Networks). Synthetic Data has lots of benefits, such as scalability, consistency, diversity and cost-effective. For scalability, synthetic data can be generated in large quantities, providing extensive datasets for training models. For consistency, synthetic data ensures consistent and accurate labeling, reducing the noise and errors associated with manual annotations. For diversity, it can introduce controlled variability, such as changes in lighting and textures, improving model robustness. For cost-effective, generating synthetic data can be more cost-effective than collecting and labeling real-world data.

In the study, there are several factors as above that could lead to better training results on this composite image. Firstly, the synthetic dataset, being larger and more diverse, likely offered the network a more comprehensive learning experience, reducing the risk of overfitting. Secondly, the impeccably labelled synthetic data ensures consistency and accuracy in training, which might not always be the case with manually annotated real-world data. Additionally, synthetic data generation can incorporate domain-specific knowledge and controlled variability, such as changes in lighting conditions, that better prepare the network for real-world scenarios. This controlled environment also allows for minimizing noise and outliers that frequently plague real-world data. Moreover, synthetic datasets can emphasize features critical for segmentation tasks, like edge clarity and texture detail, aiding the network in learning these features more effectively. It's important to note, however, that while the advantages of synthetic data are evident in our results, real-world complexity may not be entirely replicated, which can potentially limit the model's practical applicability. Thus, a combination of real and synthetic imagery could offer a more balanced and robust approach for training neural networks in tasks where both generalizability and domain-specific knowledge are crucial.

\subsection{Practical Implications}

The findings underscore the importance of synthetic data in enhancing neural network performance for image segmentation tasks.  Industries relying on accurate segmentation, such as medical imaging and autonomous driving, can benefit significantly from incorporating synthetic data into their training pipelines.

\section{Conclusion}

This study introduces a new method to generate synthetic medical images and their corresponding segmentation masks, utilizing a text-image cross-attention mechanism for training and innovation. The experimental results show that the model trained on this synthetic data is comparable to the model trained on the real data set, and in some cases superior to the model. The success of this approach offers a promising path to significantly reduce the burden of data collection and annotation in medical image analysis, with the potential to transform the field by enabling efficient, scalable, and high-quality training of segmentation models.

\bibliography{sn-bibliography}

\end{document}